\documentclass{bmvc2k}
\usepackage{booktabs} 
\usepackage{rotating}

\newcommand{\xiaosihao}{\fontsize{7pt}{\baselineskip}\selectfont}
\newcommand{\squishlist}{
	\begin{list}{$\bullet$}
		{ \setlength{\itemsep}{0pt}
			\setlength{\parsep}{1pt}
			\setlength{\topsep}{1pt}
			\setlength{\partopsep}{0pt}
			\setlength{\leftmargin}{1.5em}
			\setlength{\labelwidth}{1em}
			\setlength{\labelsep}{0.5em} } }
	\newcommand{\squishend}{
\end{list}  }

\title{Sunrise or Sunset: Selective Comparison Learning for Subtle Attribute Recognition}

\addauthor{Hong-Yu Zhou}{zhouhy@lamda.nju.edu.cn}{1}
\addauthor{Bin-Bin Gao}{gaobb@lamda.nju.edu.cn}{1}
\addauthor{Jianxin Wu}{wujx2001@nju.edu.cn}{1}

\addinstitution{
 National Key Laboratory for Novel Software Technology,\\
 Nanjing University,\\
 Nanjing, China
}

\runninghead{Zhou, Gao, Wu}{Sunrise or Sunset}

\def\etal{\emph{et al}\bmvaOneDot}

\begin{document}
	
	\maketitle
	
	\begin{abstract}
		The difficulty of image recognition has gradually increased from general category recognition to fine-grained recognition and to the recognition of some subtle attributes such as temperature and geolocation. In this paper, we try to focus on the classification between sunrise and sunset and hope to give a hint about how to tell the difference in subtle attributes. Sunrise vs. sunset is a difficult recognition task, which is challenging even for humans. Towards understanding this new problem, we first collect a new dataset made up of over one hundred webcams from different places. Since existing algorithmic methods have poor accuracy, we propose a new pairwise learning strategy to learn features from selective pairs of images. Experiments show that our approach surpasses baseline methods by a large margin and achieves better results even compared with humans. We also apply our approach to existing subtle attribute recognition problems, such as temperature estimation, and achieve state-of-the-art results.
	\end{abstract}
	
	\section{Introduction}
	\label{sec:intro}
	Recognition has been one of the central tasks in computer vision and pattern recognition, and machine learning (especially deep learning) has been one of the key forces in developing various image recognition methods and systems. The success of deep ConvNet (CNN), e.g., AlexNet~\cite{alexnet}, has greatly advanced the state of the art in image recognition.
	
	What is to be recognized in an image? The answer to this question has been constantly changing, which in consequence leads to different image recognition problems. The vision community has been recognizing properties of images with increasing difficulties: semantic image categories, fine-grained image categories and some physical attributes (such as the temperature~\cite{temperature}). Humans are good at recognizing semantic categories, while fine-grained recognition needs domain experts (e.g., bird specialists). For physical attributes, it is easy to tell whether a photo was taken at day or night while even an experienced traveler feels difficult to tell the exact place in the photo.
	
\begin{figure}
		\centering
		{\includegraphics[width=0.8\columnwidth]{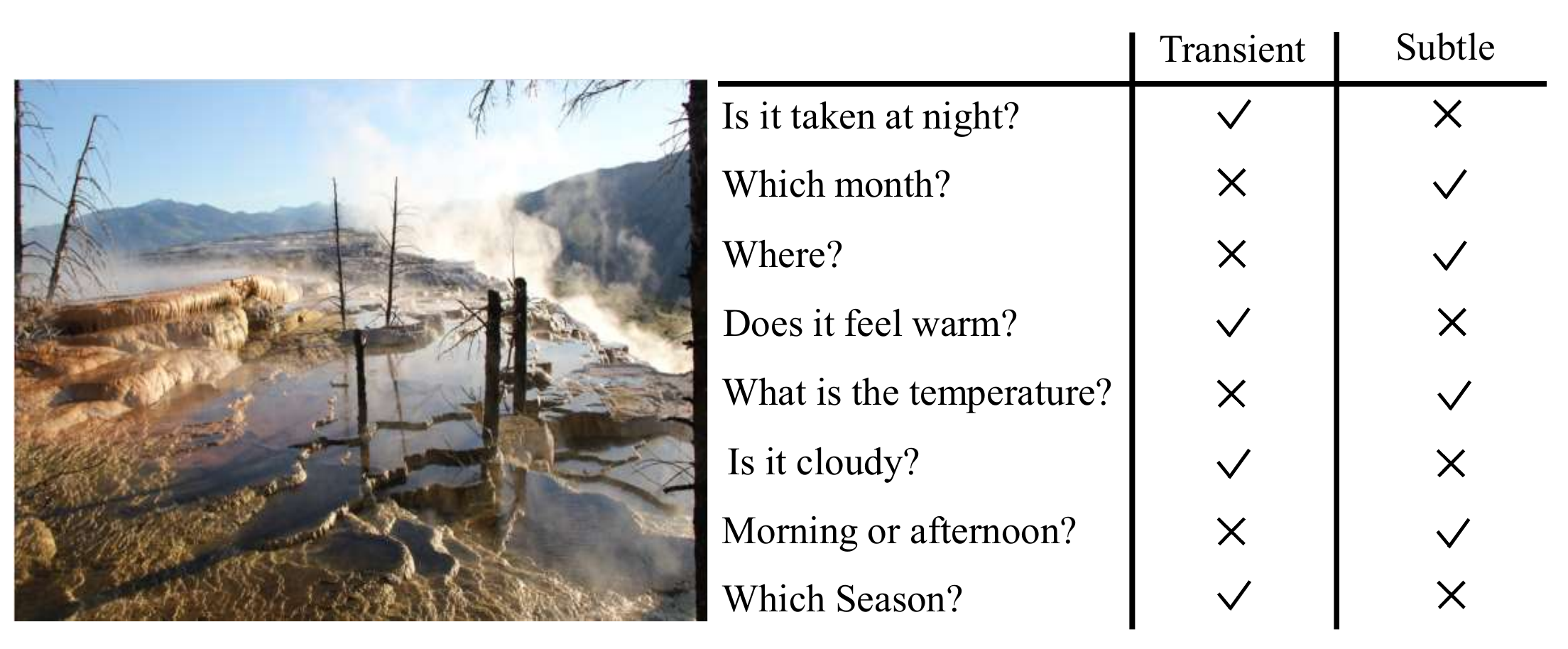}}	
		\caption{A comparison between subtle attributes and transient attributes~\cite{transient}. Compared with transient attribute, subtle attribute can be more detailed.}
		\label{comparison}
	\end{figure}

	In this paper, these seemingly indistinguishable attributes (even difficult for human beings to correctly infer from an image) are classified as {\em subtle} attributes, and we argue that it is possible to recognize this kind of attribute from images. In Figure~\ref{comparison}, we compare subtle attributes with transient attributes proposed by Laffont~\etal~\cite{transient}, which suggests that subtle attribute recognition can be harder to some extent. For simplicity, we use {\em sunrise vs. sunset}, a very difficult classification task, as an example of subtle attribute recognition and build a ``Sunrise or Sunset'' (SoS) dataset.
	
\begin{figure}
		\centering
		\subfigure[Monet's paintings]
		{\includegraphics[width=0.20\columnwidth]{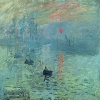} \includegraphics[width=0.20\columnwidth]{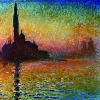} \label{fig1a}}
		\subfigure[Guess which is a sunrise?]
		{\quad\includegraphics[width=0.20\columnwidth]{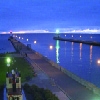} \includegraphics[width=0.20\columnwidth]{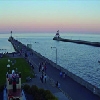} \label{fig1b}}
		\caption[examples]{Sunrise vs. Sunset: (a) Two famous paintings of Claude Monet. Left: "Impression, Sunrise"; Right: "Sunset in Venice". Notice the difference of color and luminance in these paintings. (b) Two photos taken by the same camera at the same location. One for sunrise and one for sunset. Guess which one is taken at sunrise? (answer below\footnotemark) Best viewed in color.}
		\label{figure1}
	\end{figure}

	For most people, we often marvel at the beauty of a sunrise or a sunset. In photography, the period of time witnessing sunrise or sunset is usually called the {\em golden hour} because sunrises and sunsets produce impressive photos. A probably less examined question is: What is the difference between sunrise and sunset when recorded in photos? In fact, many artists have noticed the difference and tried to recreate the difference with artistic expressions. For instance, as the founder of French Impressionist painting, Monet has composed many great paintings about sunrise and sunset by utilizing different color and luminance (cf. Figure~\ref{figure1}).
	
	\footnotetext{Answer: the photo in the right is taken at sunrise.}
	Although it might be easy to distinguish sunrise from sunset in paintings (Figure~\ref{fig1a}), it is difficult in real world photographs (Figure~\ref{fig1b}). This issue (sunrise vs. sunset) has been discussed for many years, but we still lack definitive conclusions. Some atmospheric physicists have performed considerable research cataloging and documenting the colors at both time periods. They draw a conclusion that one cannot tell the difference from the color distribution~\cite{david_william_color_light}. However, it is possible that the transition of colors (e.g., the direction along which the luminance fades) might be helpful. There are also some other factors that might affect our decision, such as lighting and shading. Human activities are also useful clues. For example, if we observe in a photo lots of cars on the road, this observation may indicate that it is more likely to be a sunset.
	
	On the SoS dataset we have built, the accuracy of sunrise vs. sunset recognition is around 50.0\% (i.e., random guess) in a group of volunteers (who are graduate students). However, after some simple training and learning, the human accuracy is improved to 64.0\%, much higher than the original performance. This fact supports that the sunrise vs. sunset distinction is (at least partially) viable. Commonly used image recognition methods (e.g., Bags of Visual Words~\cite{bovw} or CNN), however, are much worse in this task: their accuracies are between 51.7\% and 56.3\%. In other words, they are only slightly better than a random guess.
	
	In this paper, we develop a new pairwise learning strategy inspired by the different performance of human's sunrise vs. sunset recognition in various experimental setups, and then introduce a deep architecture with selective pairs of input images. Experimental results show that the proposed method (SoSNet) significantly outperforms other methods. To train and evaluate our model, we also introduce SoS, a benchmark database for discriminating sunrise from sunset images. In addition to solving the sunrise vs. sunset problem, the proposed approach is then applied to estimate the temperature given a photo, which achieved state-of-the-art experimental results too. These results suggest that our SoSNet has the potential to solve more subtle attribute recognition problems.
	
	\section{Related Work}
	\label{sec:relatedwork}
	The community has long been interested in general image recognition, i.e., recognizing objects or scenes in different semantic categories, e.g., distinguishing objects such as birds, dogs, bicycles and airplanes, or scenes such as cinema, clinic or beach. These tasks are quite easy for a human being, but not necessarily as easy for a computer. Along with the availability of larger and larger datasets (e.g., SUN~\cite{sun}, ImageNet~\cite{imagenet}), various visual representations and learning methods have been proposed to deal with general image recognition, including manually designed visual features (e.g., SIFT~\cite{sift}, SURF~\cite{surf}) and the recently popularized deep convolutional neural network learning methods. The gap between humans and computers on recognizing general objects and scenes is becoming smaller. For instance, He~\etal~\cite{kaiming} claimed that they surpass human-level performance on the ImageNet dataset for the first time. Several recent works ~\cite{locallyhybrid}~\cite{hybridcnn} also reported comparable results on the SUN 397 scene recognition dataset~\cite{sun}.
	
	Fine-grained image recognition is a more difficult recognition task, in which the objects in different categories look alike each other. Fine-grained objects in different categories only differ in some details. The CUB200-2011 dataset~\cite{cub} has 200 categories, one for each bird species. Fine-grained details are required to distinguish birds in different categories. Fine-grained recognition requires fine-grained visual representation, e.g., by finding the parts of a bird and extracting features from them rather than extracting a visual representation from the whole bird globally. Part-based methods~\cite{partmodel},~\cite{bilinear} are widely used in fine-grained recognition. Zhang~\etal~\cite{partmodel} proposed a part-based R-CNN model by utilizing part detectors. Lin~\etal~\cite{bilinear} employed a two-stream deep convolutional model to form a bilinear model which is able to produce more discriminative features.
	
	Recently, researchers also move to recognize some more difficult image attributes. Glasner~\etal~\cite{temperature} proposed a scene-specific temperature prediction algorithm that could turn a camera into a crude temperature sensor. They also made an investigation about the relation between certain regions of the image and the temperature. Weyand~\etal~\cite{planet} proposed PlaNet, a deep model to use and integrate multiple visual cues to solve the photo geolocation problem. Their deep model outperforms previous approaches and even attains superhuman levels of accuracy in some cases.
	
	Learning from pairwise input images is natural in many computer vision tasks, such as face verification~\cite{siamese}, person re-identification~\cite{reidentification}, domain adaptation~\cite{Tzeng2015SimultaneousDT} and image similarity learning~\cite{compare}. A siamese (two streams sharing the same parameters) network is the standard solution in these researches. Parikh and Grauman\cite{relative} proposed that binary attributes are an artificially restrictive way to describe images and then employed a rank function to describe some relative visual properties. Yu and Grauman~\cite{fine_grained_local_learning} used the local learning approach to resolve the problem of fine-grained visual comparisons.
	
	The proposed subtle attribute recognition can be seen as a fine-grained version of traditional attribute
	recognition and is more difficult than fine-grained image classification. We also build a SoSNet which modifies the siamese network with a ranking loss to learn representations from selective comparison.
	
	\section{Building the SoS Dataset}
	\label{section:sos_dataset}
	In this section we propose to build a SoS dataset to help us study the problem. We divide the collecting process into several stages, and individual stages are discussed in detail in the following paragraphs.
	
	{\textbf{Stage 1. Downloading images from websites.}} Modern image databases are often collected via image search engines and image hosting websites (e.g., Flickr). We cannot collect images through these channels because we cannot be sure whether those photos have been postprocessed, and their labels might be incorrect. For instance, many pictures in Flickr are labeled with both sunrise and sunset, and most of them are likely edited.
	We downloaded images from the AMOS (Archive of Many Outdoor Scenes) dataset~\cite{amos}, which contains images from nearly 30,000 webcams. For convenience, we first downloaded data between 2010 and 2016 from over 1,000 webcams and then extracted 4 images per month for each webcam.
	
	{\textbf{Stage 2. Improving data quality.}} To get exact camera locations, we only used those cameras whose uploaders attached fairly precise locations along with them. Besides, some cameras were broken while significant parts of them provided low resolution images. 
	We solved this problem by manually getting rid of broken cameras: once the webcam has one low-quality image, we delete the camera at once.
	
	{\textbf{Stage 3. Extracting locations of cameras and calculating local sunrise/sunset time.}} Note our intention is to retrieve images taken at correct \emph{local} sunrise and sunset time in order to guarantee the correctness of the whole dataset. We used IP address of each webcam to get its location, and directly employed the algorithm published by the Nautical Almanac Office to calculate local sunrise/sunset time.\footnote{\url{http://williams.best.vwh.net/sunrise_sunset_algorithm.htm}} Images captured at sunrise/sunset time were split into two classes.
	
	Finally, we obtain 12,970 images from 128 webcams over 30 countries with good quality. Note that sunrise and sunset images occupy exactly 50\% of the images in our dataset, respectively.
	Two sets of tasks are defined for SoS: an {\em easy task} and a {\em difficult task}. We randomly split the images into 10,448 training and 2,522 testing images in the {\em easy task}. For the {\em difficult task}, we use images from 104 webcams (10,448 images) for training, and the images from the rest 24 cameras for testing.
	
	\section{Selective Comparison Learning}
	We first study the performance of existing deep models and humans on the SoS dataset to assess its difficulty. The experiments that involve human also give us some hints on how to design a deep learning model to distinguish sunrise from sunset.
	\subsection{Difficulty of the SoS dataset}
	We first fine-tune the VGG-16 CNN model for the easy and difficult tasks, whose results are reported in Table~\ref{t1}. These results suggest that the fine-tuned VGG-16 CNN model has good accuracy in the easy task. In the difficult task, the accuracy (53.1\%) is close to that of a random guess. The comparison between the easy and the difficult tasks suggests that {\em a vanilla convolutional neural network might pay attention to the pixel intensities rather than the subtle attributes}. Hence, we need a different approach for the sunrise vs. sunset classification problem.
	
	{\em We also informally studied how good humans are for the difficult task}. 20 volunteers achieve roughly 52.0\% accuracy in this task when tested without preparation. After they read the answers on Quora for sunrise vs. sunset techniques, 100 images were shown to the volunteers who are required to give their predictions. In this case, the mean accuracy goes up to 64.0\%.
	
	A few interesting observations emerge in Table~\ref{t1}. Compared with hard task, VGG-16 performs much better on the easy task because the model might have memorized historical data in each webcam. On the other side, even when presented with randomly chosen single image, humans can outperform the VGG-16 model by a large margin (64.0\% vs. 53.1\%), which suggests that humans have the ability to learn to distinguish subtle attributes to some extent. Unless specified otherwise, the rest experiments are performed on the hard task which is a more realistic problem.
	\begin{table}
		\caption{Accuracy of humans and fine-tuned VGG-16 (pretrained on ImageNet) on the SoS dataset (one random image as input). mAcc means the average accuracy(\%).}
		\label{t1}
		\centering
		\begin{tabular}{c|c|ccc}
			Method    & Task & Sunrise & Sunset  & mAcc\\
			\hline
			\hline
			FT-VGG-16 & Easy & 79.6 & 80.0 & 79.8 \\
			FT-VGG-16 & Hard & 53.3 & 52.9 & 53.1 \\
			\hline
			Human    & Hard & 64.3 & 63.7 & 64.0 \\
		\end{tabular}
	\end{table}
	
	The poor performance of VGG-16 show that a single model might not be able to focus on the right parts in the input image. During human tests, we are surprised to find that given a pair of input images, especially when volunteers have been told that the pair has similar properties (e.g., the same place), they achieve much better performance. To help the recognition process, we studied the influence of different constraints, e.g., if the paired images are captured at the same day or the same place. We give 5 different constraints in Table~\ref{t2} and their test results.
	
	\begin{table}[!htp]
		\caption{Give a pair of input images, the performance of humans on the SoS dataset. Note that each volunteer is shown 5 groups of paired images corresponding to 5 different settings, and each group contains 50 pairs. {\textbf{SS}} is another restriction on each pair which requires that one image is sunrise, the other is sunset.}
		\label{t2}
		\centering
		\begin{tabular}{c|c|ccc}
			Pair Constraint & SS & Sunrise & Sunset  & mAcc\\
			\hline
			\hline
			Random pair & w/o & 65.3 & 64.9 & 65.1 \\
			Random pair & w & 67.3 & 66.9 & 67.1 \\
			\hline
			The same day & w & 67.8 & 66.8 & 67.3 \\
			The same location & w & 70.7 & 70.0 & 70.3 \\
			The same location and day & w & \textbf{72.4} & \textbf{72.4} & \textbf{72.3} \\
		\end{tabular}
	\end{table}
	
	As we can see from Table~\ref{t2}, humans perform gradually better as more constraints are added. The pure pair gets the worst result because there is too much noise. While the best result requires that both photos contain sunrise and sunset and should be taken at the same location and the same day. This phenomenon helps us to design an effective model to learn good representations.
	
	\subsection{Architecture and loss function}
\begin{figure}
		\centering
		{\includegraphics[width=0.9\columnwidth]{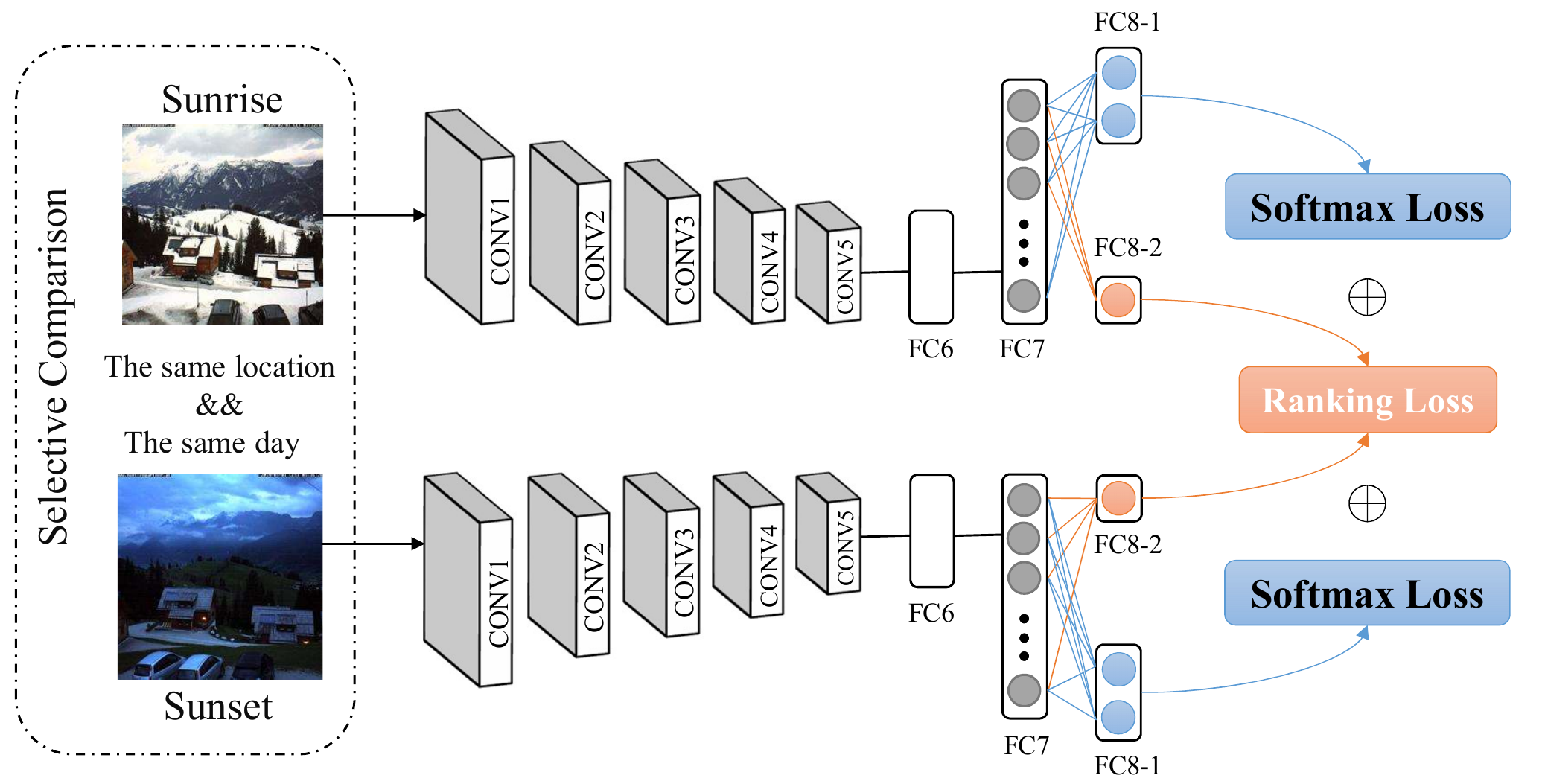}}	
		\caption{SoSNet for learning representations and prediction. This model uses the selective pairs to learn visual representations for subtle attribute recognition.}
		\label{architecture}
	\end{figure}

	Inspired by the observation, we design a two-stream network which uses a selective pair of images as its input. The two streams share the same parameters, and we denote the computation in each stream as $f_{\theta}(\cdot)$ (in which $\theta$ denotes the parameters in the two-stream model). More details about the architecture are shown in Figure~\ref{architecture}.
	
	Let us denote the two images in the pair as {\em ${X}_R$} and {\em ${X}_S$}, for sunrise and sunset images taken by the same camera in the same day, respectively. Specifically, we use $f_{\theta_1}(\cdot)$ and $f_{\theta_2}(\cdot)$ to represent the output of layer FC8-1 and layer FC8-2 in Figure~\ref{architecture}, respectively. The final loss function can be regarded as a combination of softmax loss and ranking loss,
	\begin{small}
		\begin{equation}
		\ell(X_R^{n}, X_S^{n})=\frac{1}{N}\sum_{n=1}^{N}(\ell_{softmax}(f_{\theta_1}(X_{R}^{n}),y_{R}^{n}) + \ell_{softmax}(f_{\theta_1}(X_{S}^{n}),y_{S}^{n}) + \lambda \ell_{ranking}(f_{\theta_2}(X_R^{n}),f_{\theta_2}(X_S^{n}))),
		\end{equation}
	\end{small}
	in which
	\begin{equation}
	\ell_{ranking}(f_{\theta_2}(X_R^{n}),f_{\theta_2}(X_S^{n}))=\frac{1}{1+\exp(f_{\theta_2}({X}_R^{n})-f_{\theta_2}({X}_S^{n}))},
	\end{equation}
	where $\ell_{softmax}$ is softmax loss, $N$ is the number of pairs, $\{y_R^{n}, y_S^{n}\}$ ($n=1,2,...,N$) represent image-level labels and $\lambda$ is a balancing parameter. This loss function is differentiable with respective to its parameters.
	
	During training, we use stochastic gradient descent (SGD) to minimize the average loss of all training pairs. Because we minimize the average loss between same-location same-day pairs, the pair of images $X_R$ and $X_S$ must be similar to each other in terms of the contents in them. Minimizing the ranking loss amounts to maximizing the difference between the features extracted from the pair of images using the same nonlinear mapping $f_{\theta}(\cdot)$. Hence, $f_{\theta}(\cdot)$ is forced to learn those subtle differences between the pair of images. At the same time, the softmax loss keeps the prediction correct when the input is a single image. This pairwise learning strategy discourages $f_{\theta}(\cdot)$ to capture patterns such as texture, scene or object (which vary a lot in different images).
	
	\subsection{Implementation details}
	We implemented the network in MatConvNet~\cite{matconvnet}. Both streams are initialized based on VGG-16 (pretrained on ImageNet but the last fully connected layer is removed). Then, we add three fully connected layers: FC7 (1$\times$1$\times$4096$\times$256), FC8-1 (1$\times$1$\times$256$\times$2) and FC8-2 (1$\times$1$\times$256$\times$1) to form a single stream. The balancing parameter $\lambda$ is set to 1. The batch size is 16 pairs, which means each batch actually contains 32 images. We did not use dropout or batch normalization during the training. The learning rate is $10^{-3}$ and gradually reduces to $10^{-5}$ using logspace in 20 epochs. In the test stage, we use one stream to predict the label of each input image. 
	
	\section{Experiments}
	\subsection{Experimental setup}
	In this section, we compare the accuracy of different methods on the difficult task of the SoS dataset, which is a subtle attribute recognition task. We compare the proposed approach with several baselines, ranging from classical image classification methods to recent deep convolutional models
	
	To demonstrate the effectiveness of SoSNet, we firstly consider two related methods as baselines: \textbf{Single-stream ConvNet} and \textbf{SoSNet with random pair (SoSNet-rand)}. The former one uses AlexNet~\cite{alexnet}, Network in Network (NiN)~\cite{nin}, VGG-16~\cite{vgg} net, and recent ResNet~\cite{resnet}. All models are pretrained on ImageNet. The latter baseline replaces the selective pair in SoSNet with random pair and hence make use of contrast loss~\cite{contrast} instead of ranking loss. In addition, we also compare SoSNet with the following baseline methods:
	\squishlist
	\item\textbf{Hand-crafted features + SVM.} The first family of baseline methods uses the SIFT features~\cite{sift} implemented in VLFeat~\cite{vlfeat}, which are computed at seven scales with a factor $\sqrt{2}$ between successive scales, with a stride of 4 pixels. An input image returns a set of SIFT feature vectors. We use three encoding methods to turn the set of SIFT features into one vector, including Fisher Vectors (FV)~\cite{fv}, Bags of Visual Words (BOVW)~\cite{bovw} and Vector of Locally Aggregated Descriptors (VLAD)~\cite{vlad}.
	Experiments labeled with ``aug.'' encode spatial information; ``s.p.'' use a spatial pyramid with $1\times 1$ and $3\times 1$ subdivisions. After encoding feature vectors, a linear SVM is used as a classifier.
	
	\item\textbf{Siamese features + SVM (SiameseNet).} Siamese networks have the ability to learn good features from training a similarity metric from data. Here, we implement the contrast loss function~\cite{contrast} used in person re-identification~\cite{siamese} based on VGG-16 as comparison. Let $X_i$ and $X_j$ be two training images, $c_i$ and $c_j$ be the labels for $X_i$ and $X_j$, respectively. $f_{\theta}(\cdot)$ is the deep model parameterized by $\theta$, which maps one input image to an output vector. Let $v_i=f_{\theta}(X_i)$, $v_j=f_{\theta}(X_j)$. We consider the squared Euclidean distance in the loss,
	\begin{equation}
	\ell{'}(X_i, X_j) = 
	\begin{cases}
	\| v_i-v_j \| ^2_2 & \quad \text{if } c_i = c_j \\
	\max(1-\| v_i-v_j \|^2_2,\ 0) & \quad \text{otherwise}
	\end{cases}
	\end{equation}
	In this model, the dimensionality of $v_i$ and $v_j$ is 256. We extract the last layer from siamese network as features and use a linear SVM for classification.
	\squishend
	\subsection{Experimental results}
	The sunrise vs. sunset recognition results of these baselines and the proposed method are reported in Table~\ref{t3}. In Table~\ref{t3}, the proposed method achieves an accuracy of 71.5\%, outperforming other baselines by at least 17\%. 
	
	\begin{table}[t]
		\caption{Evaluation of different methods on the hard task. All encoding methods are based on SIFT and followed by a linear SVM.}
		\label{t3}
		\centering
		\setlength{\tabcolsep}{2.5pt}
		\begin{tabular}{c|cccc|cccc|ccc}
			& \begin{turn}{65}\footnotesize FV(s.p.)\end{turn} & \begin{turn}{65}\footnotesize FV \end{turn} & \begin{turn}{65}\scriptsize VLAD(aug.) \end{turn} & \begin{turn}{65}\scriptsize BOVW(aug.) \end{turn} & \begin{turn}{65}\footnotesize AlexNet \end{turn} & \begin{turn}{65}\footnotesize NiN \end{turn} & \begin{turn}{65}\footnotesize VGG-16 \end{turn} & \begin{turn}{65}\footnotesize ResNet-101 \end{turn} & \begin{turn}{65}\footnotesize SiameseNet \end{turn} & \begin{turn}{65}\footnotesize SoSNet-rand \end{turn}& \begin{turn}{65}\footnotesize SoSNet \end{turn}\\
			\hline
			\hline
			sunrise & 54.1 & 53.4 & 50.6 & 56.6 & 52.2 & 52.1 & 53.3 & 53.8 & 54.2 & 58.6 & \textbf{70.9} \\
			sunset & 52.1 & 54.0 & 52.8 & 56.0 & 52.0 & 52.5 & 52.9 & 53.2 & 54.8  &59.0 & \textbf{71.6} \\
			mAcc & 53.1 & 53.7 & 51.7 & 56.3 & 52.1 & 52.3 & 53.1 & 53.5 & 54.5 & 58.8 & \textbf{71.2} \\
			\hline
		\end{tabular}
	\end{table}
	
	A few interesting points can be observed from Table~\ref{t3}.
	
	\emph{BOVW has higher accuracy than single-stream ConvNets.} Different from general or fine-grained image recognition, the bag of visual words models have achieved higher accuracy than single-stream ConvNets when recognizing sunrise/sunset. The drawbacks of the bag of visual words model, including small receptive field and low representational power when compared with CNN, might make it less overfit by the pixel values as the CNN.
	
	\emph{SoSNet-rand performs better than SiameseNet.} This phenomenon tells us that learning representations and predictions simultaneously (an end-to-end manner) might somehow facilitate the recognition process.
	
	\emph{SoSNet exceeds SoSNet-rand by 10 points.} Since the difference is only the pair constraint, the success suggests that selective comparison might be more useful than random pair in subtle attribute recognition.
	
	\section{Ambient Temperature Prediction}
	Glasner~\etal~\cite{temperature} first proposed to use simple scene specific temperature prediction algorithms to turn a camera into a crude temperature sensor. The data were also collected from the AMOS dataset~\cite{amos}, which mainly contains 10 different webcams. From each webcam they extracted one image every day at 11:00 am local time over a period of two consecutive years. The first year images were used for training, while the second year for testing. In this section, we follow the same experimental settings as those in~\etal~\cite{temperature}. Ten cameras are referred to as scenes (a)-(j).
	
	\begin{table*}
		\caption{Temperature estimation results for each scene in Glasner's dataset.}
		\xiaosihao
		\label{temperature}
		\setlength{\tabcolsep}{1.0pt} 
		\begin{tabular}{l| *{9}{c|}c}
			& \multicolumn{10}{c}{\footnotesize $R^2$ (the higher the better)/ $RMSE$ (the lower the better)} \\
			\hline
			\hline
			& (a) & (b) & (c) & (d) & (e) & (f) & (g) & (h) & (i) & (j) \\
			\hline
			Local Regression~\cite{temperature} & 0.67/6.85 & 0.65/7.24 & 0.70/6.03 & 0.59/4.53 & 0.76/5.77 & 0.38/3.19 & 0.50/7.63 & 0.77/5.09 & 0.10/3.68 & 0.59/7.77 \\
			\hline
			LR Temporal Win.~\cite{temperature} & 0.61/7.52 & 0.69/6.86 & 0.72/5.82 & 0.64/4.23 & 0.79/5.39 & 0.53/2.77 & 0.54/7.35 & 0.76/5.22 & 0.11/3.67 & 0.58/7.85 \\
			\hline
			Global Ridge Reg.~\cite{temperature} & 0.00/18.16 & 0.78/5.74 & 0.00/35.02 & 0.00/11.37 & 0.00/43.51 & 0.10/3.84 & 0.74/5.54 & 0.00/13.86 & 0.23/3.41 & 0.46/8.91 \\
			\hline
			CNN~\cite{temperature} & 0.49/8.55 & 0.79/5.59 & 0.71/5.96 & 0.24/6.17 & 0.61/7.30 & 0.48/2.90 & 0.39/8.48 & 0.79/4.88 & 0.43/2.93 & 0.66/7.12\\
			\hline
			Transient Attrib.~\cite{transient} & 0.36/9.60 & 0.70/6.69 & 0.58/7.20 & 0.55/4.75 & 0.68/6.62 & 0.21/3.59 & 0.58/7.03 & 0.65/6.31 & 0.16/3.56 & 0.67/7.00\\
			\hline
			FC6~\cite{hon2} & 0.52/8.28 & 0.80/5.46 & 0.61/6.89 & 0.56/4.72 & 0.80/5.30 & 0.21/3.60 &0.54/7.34 & 0.79/4.90 & 0.06/3.78 & 0.59/7.80 \\
			\hline
			Pool4~\cite{hon2} & 0.58/7.79 & 0.84/4.87 &  0.79/5.03 & 0.60/4.45 & 0.87/4.22 & 0.40/3.14 & 0.63/6.61 & 0.80/4.72 & 0.52/2.70 & 0.76/6.01 \\
			\hline
			Ours & \textbf{0.73/6.26} & \textbf{0.89/4.57} & \textbf{0.83/4.92} & \textbf{0.70/3.80} & \textbf{0.90/3.98} & \textbf{0.58/2.53} & \textbf{0.80/5.20} & \textbf{0.86/3.95} &\textbf{0.55/2.48}& \textbf{0.78/5.81} \\
			\hline
		\end{tabular}
	\end{table*} 
	
	\subsection{Baseline methods and evaluation metric}
	Glasner~\etal~\cite{temperature} described 5 different estimation methods: Local Regression (LR), Local Regression with a Temporal Window (LRTW), Global Regularized Regression (GRR), CNN and Transient Image Attributes (TA). The first three use simple pixel intensities as features while the last two use more sophisticated global image features.
	
	There are two protocols to evaluate the performance of aforementioned algorithms. The coefficient of determination ($R^2$) and Root Mean Squared Error (RMSE).
	Glasner~\etal~\cite{temperature} used $R^2$ to compare results for different scenes, while RMSE provides an intuitive interpretation of the estimate's quality. More details of these two protocols can be found in~\cite{temperature}.
	
	\subsection{Temperature prediction with selective comparison}
	Unlike the experiments in sunrise/sunset, {\em the selective pairs are restricted to photos taken at the same place}. And we replace the softmax loss with square loss to predict the temperature. We train an independent SoSNet for each scene. The learning rate is $10^{-3}$ and gradually reduces to $10^{-5}$ using logspace in 10 epochs. Then we follow the same instructions in~\cite{hon2} which extracts Pool4 as features for a linear $\nu$-SVR using LIBSVM. For fairness, we set the SVR parameters as $C = 100$, $\nu = 0.5$ and $g=1$ for all our experiments which are the same as those in~\cite{temperature}. More details can be found in~\cite{temperature} and~\cite{hon2}.
	
	We report the performance of our approach and seven other algorithms on ten different scenes in Table~\ref{temperature}. Our method achieves the best results on all 10 scenes. It is worth noting that our approach outperforms the traditional CNN by a large margin.
	
	\section{Conclusion}
	In this paper, we have proposed to study a new type of image recognition problem: to recognize subtle attributes. We built a SoS dataset, and our experiments showed that both humans and existing computer vision and machine learning methods have poor accuracy on this dataset. We proposed a model named SoSNet that learns discriminative features using selective pairs as inputs while is still able to make predictions. The same SoSNet also achieved state-of-the-art results in temperature prediction from an image.
	
	\section*{Acknowledgements}
	This work was supported in part by the National Natural Science Foundation of China under Grant No. 61422203. 
	\bibliography{egbib}
\end{document}